\pdfoutput=1
\documentclass[runningheads]{llncs}
\usepackage{graphicx}

\usepackage{amsfonts}
\usepackage{amsmath}
\usepackage{multirow}
\usepackage{booktabs}
\usepackage{tablefootnote}
\usepackage{threeparttable}
\usepackage{color} 
\usepackage{hyperref}
\hypersetup{
    colorlinks=true,
    linkcolor=black,
    filecolor=black,      
    urlcolor=black,
    citecolor=black,
}
\usepackage[all]{hypcap}
\usepackage{bbding}

\begin{document}
\title{Inheritance-guided Hierarchical Assignment for Clinical Automatic Diagnosis}
\titlerunning{Inheritance-guided Hierarchical Assignment for Clinical Automatic Diagnosis}
%
\author{Yichao Du\inst{1} 
\and Pengfei Luo\inst{1}
\and Xudong Hong\inst{2}
\and Tong Xu\inst{1(}\Envelope\inst{)}
\and Zhe Zhang\inst{1}
\and Chao Ren\inst{1}
\and Yi Zheng\inst{3}
\and Enhong Chen\inst{1}
}
\authorrunning{Y. Du et al.}
%
\institute{School of Computer Science and Technology, University of Science and Technology of China, Hefei, China\\
\email{\{duyichao,pfluo,renchao\}@mail.ustc.edu.cn, zhezhang1527@163.com, \{tongxu,cheneh\}@ustc.edu.cn}
\and Institution of Smart City Research (WuHu), University of Science and Technology of China, Wuhu, China\\
\email{xdhong@ahut.edu.cn} \and
HUAWEI Technologies\\
\email{zhengyi29@huawei.com}}
\maketitle              
\begin{abstract}
Clinical diagnosis, which aims to assign diagnosis codes for a patient based on the clinical note, plays an essential role in clinical decision-making. Considering that manual diagnosis could be error-prone and time-consuming, many intelligent approaches based on clinical text mining have been proposed to perform automatic diagnosis. However, these methods may not achieve satisfactory results due to the following challenges. First, most of the diagnosis codes are rare, and the distribution is extremely unbalanced.  Second, existing methods are challenging to capture the correlation between diagnosis codes. Third, the lengthy clinical note leads to the excessive dispersion of key information related to codes.  To tackle these challenges, we propose a novel framework to combine the inheritance-guided hierarchical assignment and co-occurrence graph propagation for clinical automatic diagnosis. Specifically, we propose a hierarchical joint prediction strategy to address the challenge of unbalanced codes distribution. Then, we utilize graph convolutional neural networks to obtain the correlation and semantic representations of medical ontology. Furthermore, we introduce multi attention mechanisms to extract crucial information. Finally, extensive experiments on MIMIC-III dataset clearly validate the effectiveness of our method.

\end{abstract}

\section{Introduction}
The clinical note is an essential part of Electronic Health Record (EHR), 
which contains lengthy and terminological 
text records about medical history, chief complaint, current symptoms, and laboratory test results. 
To avoid the redundancy and ambiguity caused by the text, 
the World Health Organization recommends using the diagnosis codes in the
International Classification of Diseases (ICD) for each disease, 
symptom, and sign to represent the patient's condition.
The goal of clinical diagnosis is to assign the most likely diagnosis codes for the patient based on the clinical note. 
Traditionally, clinical diagnosis is completed by well-trained clinical coders, which is labor-intensive and error-prone because the diagnosis codes system is vast and growing.
For example, in the United States, about 20\% of patients are misdiagnosed at the primary care level, 
and one-third of the misdiagnosis will cause later severe injury to the patients~\cite{singh2017global}.

\begin{figure} 
    \label{intro_fig1}
    \centering
    \includegraphics[width=\textwidth]{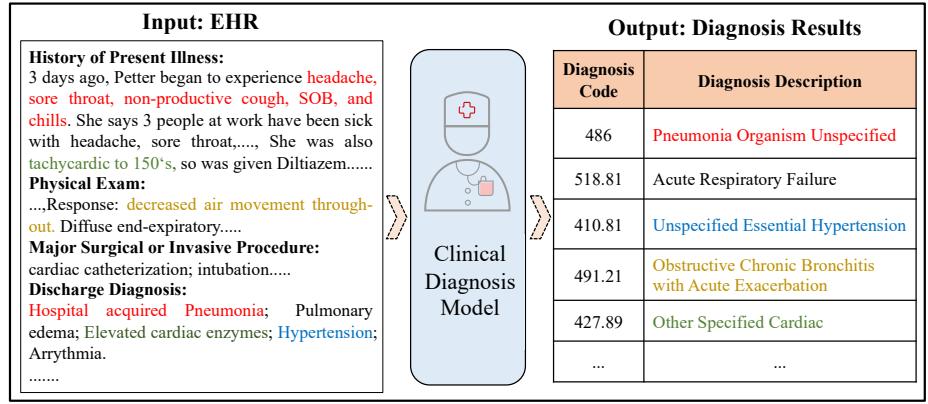}
    \caption{Illustration of clinical automatic diagnosis task. The input and output of the model are EHR and diagnosis codes, respectively. The text related to the diagnosis code in the EHR is marked in colored font.} \label{intro_fig1}

\end{figure}

Consequently, the automatic clinical diagnosis based on EHR has aroused widespread attention in the industrial and academic circles~\cite{esteva2019guide}.
Among the proposed methods, supervised machine learning methods were trained to learn shallow feature combinations 
for clinical note~\cite{perotte2014diagnosis,kavuluru2015empirical}.
Recently, most deep learning models treated this task as a sequence learning problem, 
including used Convolutional Neural Networks~\cite{mullenbach2018explainable,li2020icd} and Recurrent Neural Networks~\cite{shi2017towards,baumel2017multi} to capture complex semantic information. 
On this basis, medical ontology was further introduced as auxiliary knowledge.
Specifically, Bai et al.~\cite{bai2019improving} incorporated the disease encyclopedia of Wikipedia into the model to enhance its predictive ability. 
Besides, the patient's history and demographic information could also be leveraged to enhance the prediction of future admissions~\cite{prakash2016condensed,bai2019improving,ma2018kame}. 
Although these methods have made significant progress in automatic diagnosis, they may also fail due to the following challenges:

\begin{itemize}
    \item \textbf{C1: The number of diagnosis codes is enormous, and the distribution is extremely unbalanced.} 
    For example, the MIMIC-III~\cite{johnson2016mimic} dataset, which is widely used for automatic diagnosis, contains 8,925 codes, 
    but 4,344 appear less than five times in all data.
    The severe long-tail distribution makes it difficult to assign proper codes to rare diseases, which may cause irreparable damage to the patients.
    \item \textbf{C2: The correlations between diagnosis codes are greatly overlooked.}
    However, the medical relationship between diseases can help us identify diseases 
    that are not clearly reflected by the clinical note.
    As shown in Fig. \ref{intro_fig1}, we can extract clues (colored fonts) from the text 
    to assign diagnosis codes to the patient. 
    For example, from the text ``Hospital Acquired Pneumonia", we can easily infer the code ``486 (Pneumonia Organism Unspecified)". 
    Nevertheless, it is difficult to infer the code ``410.81 (Acute Respiratory Failure)" only from the text. 
    Fortunately, we can infer the code ``410.81'' from the relationship between it and the code ``486'',
    that is, ``Pneumonia Organism Unspecified" will in all probability cause patients to have 
    the symptom of ``Acute Respiratory Failure".
    \item \textbf{C3: In clinical note, only a few key fragments can provide valuable information for automatic diagnosis.} 
    For example, in the MIMIC-III dataset, clinical notes usually contain more than 1,500 tokens, 
    but only a few tokens are related to specific diagnosis codes.
    Extracting crucial tokens for specific diagnosis codes is as tricky as finding a needle in a haystack. 
\end{itemize}

To this end, 
we propose a model named \textbf{I}nheritance-guided \textbf{H}ierarchical Assignment with \textbf{C}o-occurrence-based \textbf{E}nhancement (IHCE) to address these challenges.
First, for C1, we design a hierarchical assignment method based on the hierarchical inheritance structure of diagnosis codes defined by ICD, which makes assignment level by level. 
As shown in Fig.~\ref{intro_fig2}, ``405.0 (Malignant renovascular hypertension)'' and ``405.1 (Benign secondary hypertension)'' are mutually exclusive. 
Moreover, ``405.01 (Malignant renovascular hypertension)'' inherits the information of ``405.0''. 
Consequently, if we assign ``405.0'' at the high level, we will tend to further assign ``405.01'' instead of the children of ``405.1''. 
With the inheritance-guided hierarchical assignment, we can use the diagnostic results of a high level to guide the low level, which addresses the challenge of unbalanced distribution. 
Second, for C2, we construct a co-occurrence graph based on EHR data 
and use GCN to obtain the diagnosis codes' semantic representations.
In this way, the representations of the diagnosis codes contain the correlation between diseases, which help us to assign codes to diseases for where it is challenging to find textual clues from the clinical note.
Third, for C3, we enhance the ability to extract the tokens related to the 
diagnosis codes based on the attention mechanism which models the interaction between diagnosis codes' ontology representations and the clinical note.
Finally, experiments on a real medical dataset show that IHCE is superior to the 
SOTA methods on all evaluation metrics.
\begin{figure}
    \label{intro_fig2}
    \centering 
    
    \includegraphics[scale=0.35]{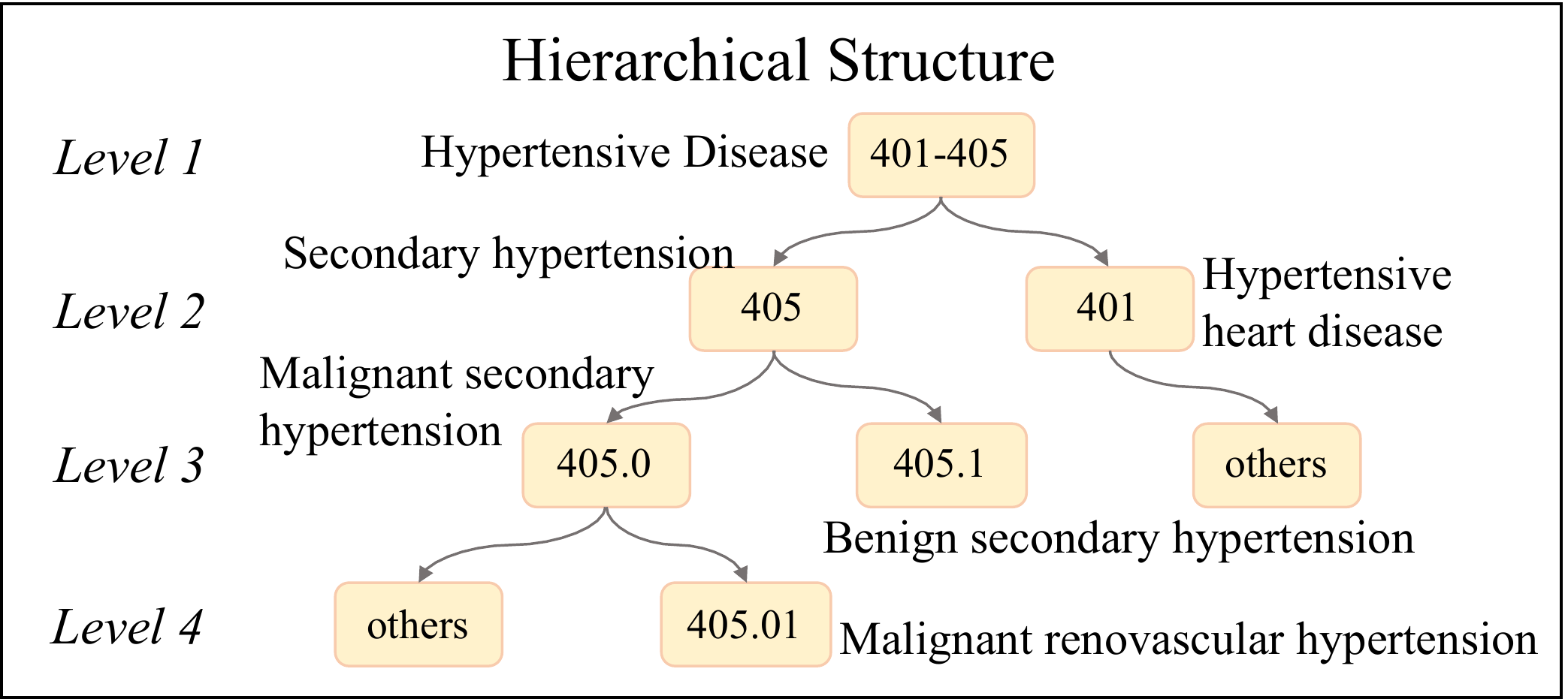}
    \caption{An example of diagnosis codes' descriptors and their hierarchical inheritance structure based on ICD.} \label{intro_fig2}
    
\end{figure}

\section{Related Work}
\subsection{Clinical Automatic Diagnosis}
Clinical automatic diagnosis has become a research hot spot in medicine, aiming to solve manual diagnosis limitations.
In recent years, deep learning technologies~\cite{shi2017towards,mullenbach2018explainable,li2020icd} have shown substantial advantages over traditional machine learning methods~\cite{perotte2014diagnosis,kavuluru2015empirical} 
and have been widely used for this task. 
Most researchers modeled this task as a multi-label text classification task based on the free text in EHR. 
Among them, Shi et al.~\cite{shi2017towards} proposed a character-perceived LSTM network that generated written diagnosis descriptions 
and representations of diagnosis codes. 
Baumel et al.~\cite{baumel2017multi} proposed a hierarchical-GRU with a label-dependent 
attention layer to alleviate excessive text problem.
Wang et al.~\cite{wang2018joint} proposed a label-word joint embedding model and applied the cosine similarity to assign the codes.
Moreover, some researchers incorporated external knowledge into the model~\cite{prakash2016condensed,bai2019improving,ma2018kame}. 
For example, Knowledge Source Integration (KSI)~\cite{bai2019improving} calculated the matching score between the clinical note and 
each knowledge document based on the intersection of clinical notes 
and external knowledge for this task.
Our method is different from these methods, considering the hierarchy and co-occurrence relationship to achieve better performance in automatic diagnosis.

\subsection{Graph Convolutional Network}
In the past few years, Graph Convolutional Network (GCN)~\cite{kipf2016semi} has been widely used in various tasks to encode advanced graph structures,
such as healthcare~\cite{yichao6automatic,liu2020coupled}, recommender systems~\cite{liu2020exploiting}, 
business analysis~\cite{li2020competitive}, machine translation~\cite{bastings2017graph}, text classification~\cite{yao2019graph,peng2018large}.
Specifically, in order to promote the sharing of disease among patients, 
Liu et al.~\cite{liu2020coupled} applied GCN on text corpus to collect high-order neighbor information, 
and predicted for patients based on projection.
Yao et al.~\cite{yao2019graph} proposed Text-GCN, 
which was utilized to learn the representations of words and documents to improve text classification.
Peng et al.~\cite{peng2018large} proposed a recursive regularized GCN to perform large-scale text classification on word co-occurrence graphs.
Inspired by this, we apply GCN to obtain a good correlation between diagnosis codes and represent the medical ontology.
Furthermore, we utilize the ontology representations as interactive information to improve the performance of automatic diagnosis.

\section{Preliminaries}
For a patient, the word sequence $S=\left\{w_1,w_2,...,w_n\right\}$ of the patient's clinical note is included, where $n$ is the length of $S$. 
Furthermore, a set of diagnosis codes   $ L=\left\{l_1, l_2,..., l_{|L|} \right\} \in \left\{0,1\right\}^{|{{L}}|}$ are also contained to denote the diseases of the patient, where $|L|$ is the number of diagnosis codes. 
In addition, we also introduce hierarchical inheritance structure $\mathcal{L}=\left\{L^1,L^2,...,L^\mathcal{T} \right\}$  to expand $L$ based on external knowledge (i.e., the hierarchical inheritance structure based on ICD in Fig.~\ref{intro_fig2}), where  $L^t=\left\{l^t_1, l^t_2,...,l^t_{|L^t|} \right\}$ means all diagnosis codes of the level-$t$, and $\mathcal{T}$ is the total number of hierarchical levels. 
Note that,
$L^\mathcal{T}=L$, which means that the last hierarchical level is the same as the patient's diagnosis codes.
With above description, we can define the clinical automatic diagnosis task with inheritance guidance as follows:
\begin{definition}
Given the patient's clinical note sequence $S$ and the diagnosis codes hierarchical inheritance structure $\mathcal{L}$, our goal is to predict the patient's diagnosis codes set $\hat{{L^t}}=\left\{\hat{{l^t_1}}, \hat{l^t_2},... \right\}\in \left\{0,1\right\}^{|\hat{{L^t}}|}$ level by level, and finally use the last level $\hat{L^{\mathcal{T}}}$as the prediction of the patient's diagnosis.
\end{definition}

\section{The Proposed Model IHCE}
\begin{figure}
    \includegraphics[width=\textwidth]{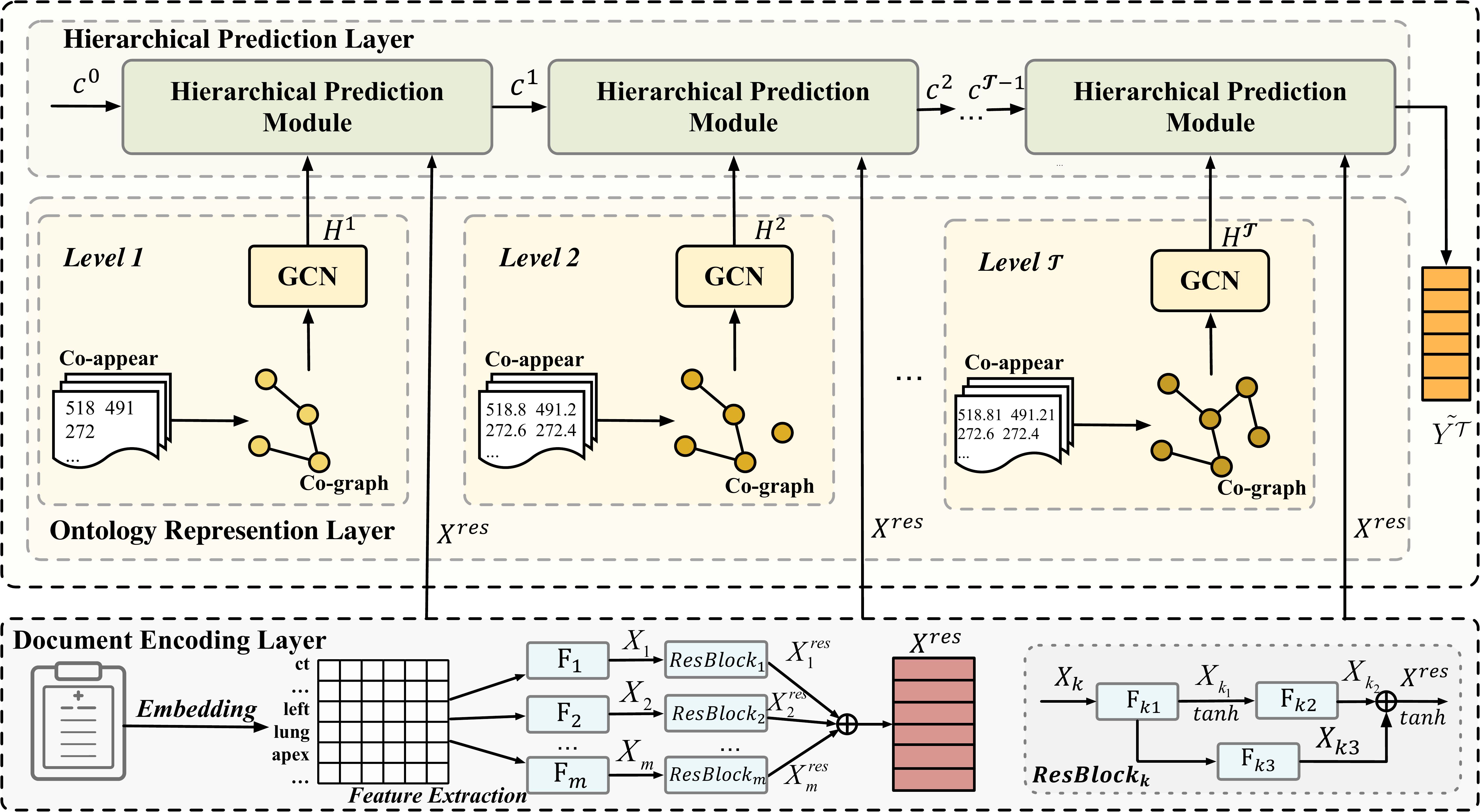}
    \centering
    \caption{The architecture of IHCE.} \label{model_fig1}
\end{figure}

As shown in Fig.~\ref{model_fig1}, IHCE mainly contains three components: (1) Document Encoding Layer (DEL), 
(2) Ontology Representation Layer (ORL), and (3) Hierarchical Prediction Layer (HPL).
Specifically, we first utilize the DEL to obtain representations of the clinical note and diagnosis codes. 
Secondly, we apply the ORL to obtain the correlation and semantic representations of medical ontology. 
Finally, we design HPL to predict the patient’s diagnosis codes based on hierarchical dependence and attention mechanism.

\subsection{Document Encoding Layer}
The goal of DEL is to generate unified representations for the clinical note and diagnosis codes. 
We first utilize the Embedding Module to encode the patient's clinical note and diagnosis codes. 
Then, we apply the Feature Extraction Module to enhance the semantic representation of the clinical note.

\subsubsection{Embedding Module.}
First, given the word sequence $S=\left\{w_1,w_2,...,w_n\right\}$, 
we use the word vector matrix $E=\left[e_{1}, e_{2}, \ldots, e_{|E|}\right] \in \mathbb{R}^{|E| \times d_{e}}$ to obtain the word embedding sequence $X=\left[x_{1}, x_{2}, \ldots, x_{n}\right] \in \mathbb{R}^{n \times d_{e}}$, 
where $|E|$ is the size of the vocabulary, and $d_e$ is the dimension of the word vector.
Similarly, we generate the diagnosis code ontology embedding for each code $l^t_i\in L^t$ via averaging the word embedding of its descriptor sequence:
\begin{equation}
    \label{eq1}
    \begin{array}{l}
        v_{i}^t=\frac{1}{\left|N_i^t\right|} \sum_{j \in N_i^t} e_{j}, \quad i=1, \ldots,|L^t| \\
        V^t=\left[v_{1}^t, v_{2}^t, \ldots, v_{|L^t|}^t\right] \in \mathbb{R}^{|L^t| \times d_{e}}
    \end{array},
\end{equation}
where $N_i^t$ is the  text descriptor index set of  $l_i^t$, 
and $v_i^t$ denotes the word embedding of the $l_i^t$,
and $V^t$ indicates the representations of all codes of the level-$t$.

\subsubsection{Feature Extraction  Module.}
As shown in the lower part of the Fig.~\ref{model_fig1}, we apply the 
multi-filter residual convolutional neural network~\cite{li2020icd} architecture 
for deep feature extraction on clinical note's embedding matrix $X$.

First, we utilize convolutional neural networks containing $m$ filters to capture different length patterns of word sequence:
\begin{equation}
    \label{eq2}
    \begin{array}{c}
        X_{1}=\mathrm{F}_{1}\left(X, W_{1}\right)=\tanh \left[\ldots, W_{1}^{T} X^{j: j+s_{1}-1}, \ldots\right] \\
        \dots \\
        X_{m}=\mathrm{F}_{m}\left(X, W_{m}\right)=\tanh \left[ \ldots, W_{m}^{T} X^{j: j+s_{m}-1}, \ldots\right]
    \end{array}, \text{where }j=1,2,...,n,
\end{equation}

Let us take the $k$-th operation as an example. 
$\mathrm{F}_{k}\left(X, W_{k}\right)$ denotes the convolution operation on the matrix $X$, 
where $W_{k} \in \mathbb{R}^{\left(s_{k} \times d_{e}\right) \times d_{c}}$ is the parameter matrix, 
and $d_c$ indicates each convolutional layer's feature mapping dimension. 
$s_1,s_2,...,s_m$ denote different convolution kernel sizes, 
and $X^{j: j+s_{k}-1} \in \mathbb{R}^{s_{k} \times d_{e}}$ is the input matrix of the $j$-th to the $(j+s_k-1)$-th rows in $X$. 
Note that, we set padding and stride as $floor(s_k/2)$ and $1$.
Finally, the feature matrices $X_k \in \mathbb{R}^{n \times d_c}, k=1,2,...,m$ can be obtained. 
In order to express conciseness, the bias is ignored in all the calculation formulas in this paper.

Next, we connect $m$ parallel residual blocks after the multi-filter convolutional layer, 
capturing longer text features by expanding the receptive field. 
Taking the $k$-th unit as an example, the residual block is formally defined as:
\begin{equation}
    \label{eq3}
    \begin{array}{c}
        X_{k_{1}}=\mathrm{F}_{k_{1}}\left(X_{k}, W_{k_{1}}\right)=\tanh \left[ \ldots, W_{k_{1}}^{T} X_{k}^{j: j+s_{k}-1}, \ldots\right], \\
        X_{k_{2}}=\mathrm{F}_{k_{2}}\left(X_{k_{1}}, W_{k_{2}}\right)=\left[ \ldots, W_{k_{2}}^{T} X_{k_{1}}^{j: j+s_{k}-1}, \ldots, \right],\\
        X_{k_{3}}=\mathrm{F}_{k_{3}}\left(X_{k}, W_{k_{3}}\right)=\left[\ldots, W_{k_{3}}^{T} X_{k}^{j: j}, \ldots\right], \\
        X_{k}^{r e s}=\tanh \left(X_{k_{2}}+X_{k_{3}}\right),
    \end{array}
\end{equation}
where $j=1,2,...,n$, and $W_{k_i}$ is the weight matrix of the $k_i$-th convolution layer in the 
residual block, specifically  $W_{k_{1}} \in \mathbb{R}^{\left(s_{k} \times d_{c}\right) \times d_{r}},  W_{k_{2}} \in \mathbb{R}^{\left(s_{k} \times d_{r}\right) \times d_r}, W_{k_{3}} \in\mathbb{R}^{\left(1 \times d_{c}\right) \times d_r}$. 
The output of each residual block is $X_{k}^{r e s},k=1,2,...,m$, 
where $d_r$ indicates the feature mapping dimension. 
Finally, we concatenate them together by rows to obtain an enhanced clinical note's representation:
\begin{equation}
    \label{eq4}
    X^{r e s}=\operatorname{concat}
    \left(X_{1}^{r e s}, \ldots, X_{m}^{r e s}\right) \in 
    \mathbb{R}^{n \times d_{res}}, \text{where } d_{r e s} = {\left(m \times d_{r}\right)}.
\end{equation}

\subsection{Ontology Representation Layer}
Comorbidities and complications manifest the correlation between the diagnosis codes ontology and play an auxiliary role for codes that are difficult to predict based on the clinical note alone.
To this end, we first use co-occurrence features at each hierarchical level to construct a co-occurrence graph (co-graph) of diagnosis codes ontology.
Then, we use GCN to capture the ontology's representations, which contain the correlation between the ontology.
Here we take the level-$t$ as an example to introduce the process.

\subsubsection{Co-graph Construction.}
The co-graph is represented by $G^t=\left(L^t,E^t\right)$, where $L^t$ and $E^t$ indicate the diagnosis codes set and edge set of the level-$t$, respectively.
For any diagnosis code $l^t_i$, if there is another code $l^t_j$ in the EHR data that co-appears, 
there is an edge $e(l^t_i,l^t_j)$ between them. And the corresponding weight is calculated as follows:
\begin{equation}
    \label{eq5}
    e(l^t_i,l^t_j)=\frac{\mathrm{count}(l^t_i, l^t_j)}{\sum_{l^t_k \in L^t}{\mathrm{count}(l^t_i, l^t_k)}},
\end{equation}
where $\mathrm{count}(\cdot,\cdot)$ indicates the number of times the two codes co-appear  in the whole EHR dataset, 
which can represent prior knowledge. 
After that, the edge set $E^t$ can be described as follows:
\begin{equation}
    \label{eq6}
    E^{t}=\{ e(l^t_i,l^t_j) \mid l^t_i, l^t_j \in L^t\}.
\end{equation}

\subsubsection{Co-graph Propagation via GCN.} 
Now we turn to represent the diagnosis codes.
First, we can obtain the feature matrix $H^{t,\left(0 \right)}=V^t \in \mathbb{R}^{|L^t| \times d_e}$ 
of the diagnosis codes ontology by Equation (\ref{eq1}). 
For the sake of simplicity, we omit the superscript $t$ in the rest of this subsection.
Then, we apply the GCN to propagate the representations of the diagnosis codes on the co-graph $G$, 
which takes the feature matrix $H^{(l)}$ and the matrix $\tilde{A}$ as input, 
and update the embedding of the codes by utilizing the information of adjacent codes:
\begin{equation}
    \label{eq7}
    H^{(l+1)}=\sigma\left(\tilde{D}^{-\frac{1}{2}} \tilde{A} \tilde{D}^{-\frac{1}{2}} H^{(l)} W^{(l)}\right),
\end{equation}
where $\tilde{A}=A+I$, $A$ is the adjacency matrix of $G$, $I$ is the identity matrix,
$\tilde{D}_{i i}=\sum_{i} \tilde{A}_{i j}$,
and $W^{(l)}$ is a layer-specific trainable weight matrix.
$\sigma(\cdot)$ denotes an activation function, such as the $\mathrm{ReLU}(\cdot)= \mathrm{max}(0,\cdot)$.
$H^{(l)} \in \mathbb{R}^{L \times d_{g}}$ is the matrix of activations in the $l$-th layer, where $d_g$ indicates the hidden layer size of GCN. Then the last hidden layer is used to represent the diagnosis codes ontology,
i.e., $H^{t}=H^{t,\left(l+1 \right)} \in \mathbb{R}^{|L^t| \times d_g}$.

\subsection{Hierarchical Prediction Layer}
To simulate human diagnosis's gradual progress from shallow to deep, 
we propose an inheritance-guided hierarchical joint learning mechanism. 
To be specific, according to the hierarchical structure of the codes, the patient is diagnosed progressively from coarse-grained to fine-grained.

\begin{figure}
    \centering 
    \includegraphics[scale=0.35]{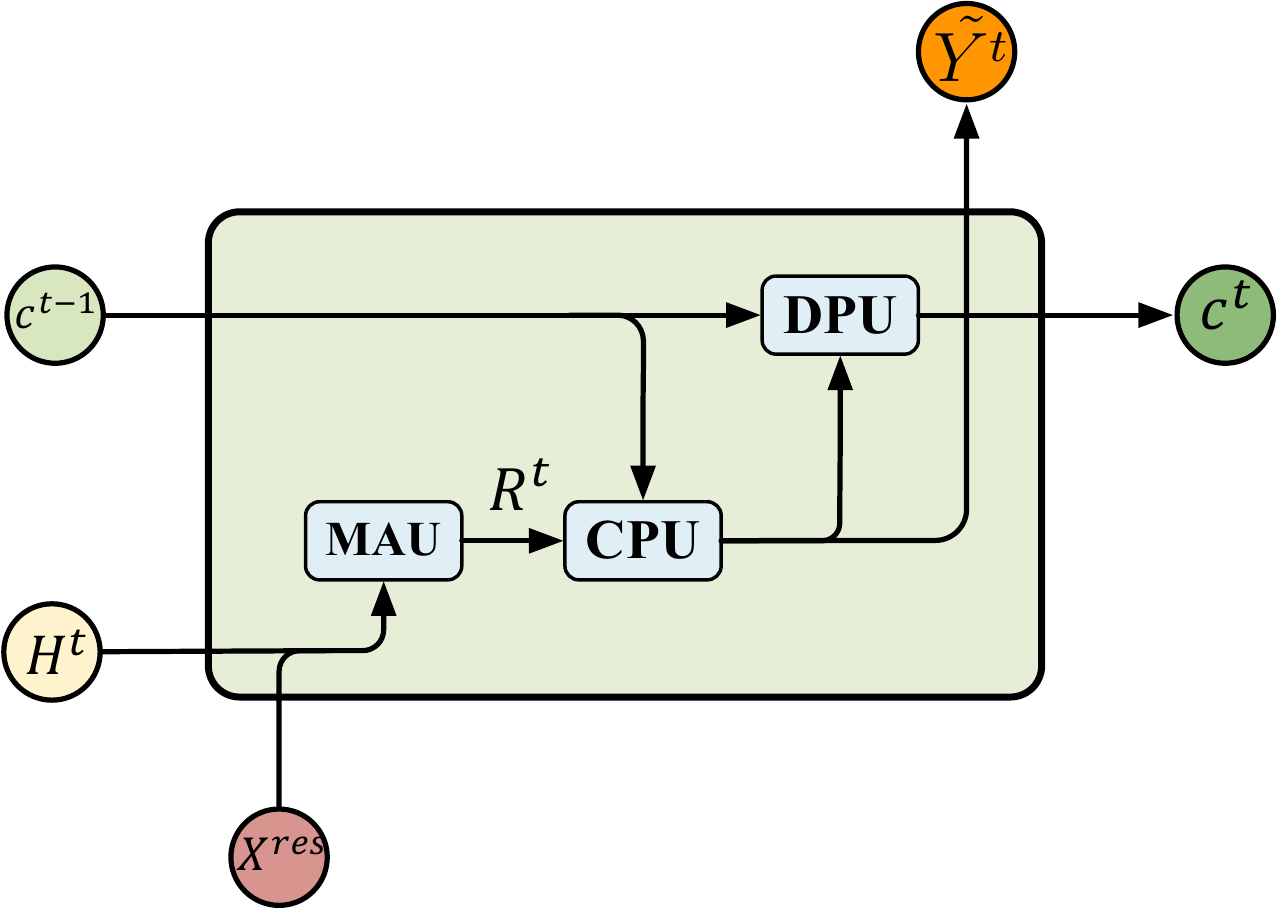}
    \caption{Hierarchical Prediction Module.} \label{model_fig2}
\end{figure}

Fig.~\ref{model_fig2} shows the core module Hierarchical Prediction Module(HPM) of HPL.
Specifically, HPM is mainly composed of three parts, 
namely Multi Attention Unit (MAU), Code Predicting Unit (CPU) and Dependency Passing Unit (DPU) respectively.
For the level-$t$, the input of HPM includes three parts, i.e.,
the clinical note's representation $X^{res}$, 
the medical ontology representations $H^{t}$, 
and the dependency information $c^{t-1}$ of the previous level:
\begin{equation}
    \label{eq8}
    \begin{array}{l}
    R^{t}=\mathrm{MAU}\left(X^{res}, H^{t}\right), \\
    Y^{t}=\mathrm{CPU}\left(c^{t-1}, R^{t}\right), \\
    c^{t}=\mathrm{DPU}\left(c^{t-1}, \tilde{Y^{t}}\right).
    \end{array}
\end{equation}
We first utilize the MAU part to obtain the correlation representation $R^{t}$ between the clinical note and medical ontology.
Next, the CPU part assigns the diagnosis codes $\tilde{Y^{t}}$ to the patient based on the $R^{t}$ and $c^{t-1}$. 
Finally, the DPU part generates the level dependency information $c^{t}$ for the next level based on the previous level's memory and the current level's assignment results.
Note that we set $c^0$ to $0$ since the current level is $0$ and does not contain the previous level's information. Next, we introduce each unit of the HPM at level-$t$.

\subsubsection{Multi Attention Unit.}
By the operations above, we can obtain the clinical note representation $\boldsymbol{}{X}^{res}$ 
and medical ontology representations $H^t $.
Intuitively, the patient's clinical note is composed of a large number of lengthy text descriptions 
and different codes may focus on different aspects of the document.
Therefore, for level-$t$, we need $|L^t|$ aspects to focus on different codes to represent the overall semantic of the whole clinical note. 
Next, we introduce the two attention mechanisms we use.
\paragraph{Ontology Guided Attention.}
For some diagnosis codes that are difficult to predict using only clinical text, 
we can improve it by interacting between the clinical note and medical ontology.
First, we pass the document feature matrix $X^{res}$ through a simple feed-forward neural network:
\begin{equation}
    \label{eq9}
    O_t'=\tanh (W_t' \cdot ({X^{res}})^T), \\
\end{equation}
where $W_t' \in\mathbb{R}^{d_g \times d_{res}} $ is the transform matrix, $d_g$ is consistent with the dimension of the columns of $H^t$, 
and $O_t' \in \mathbb{R}^{d_g \times n}$ is the intermediate result.
Then, for each code $l^t\in L^t$, 
we can generate the attention vector guided by the ontology:
\begin{equation}
    \alpha_{l^t}=\operatorname{softmax}(h_{l^t} \cdot O_t'),
\end{equation}
where $h_{l^t} \in H^t$ is the feature vector of label $l^t$, 
and $\mathrm{softmax}(\cdot)$ is the normalized exponential function for row operations.
The attention $\mathbf{\alpha}_{l^t} \in \mathbb{R}^{1 \times n}$ is then used to compute vector representation for each label:
\begin{equation}
    {x^{att}_i}' = \alpha_{l^t} \cdot X^{res},
\end{equation}

Finally, we concatenate the ${x_i^{att}}'(i=1,..,|L^t|)$ to obtain the ontology guided document representation, 
denoted as ${X_t^{att}}'=[{x^{att}_1}',{x^{att}_2}',...,{x^{att}_{|L^t|}}']\in \mathbb{R}^{|L^t|\times d_{res}}$.

\paragraph{Code Specific Attention.}
Similar to ontology guided attention, the code specific attention is formalized as:
\begin{equation}
    \label{eq12}
    \begin{array}{l}
        O_t'' =\tanh (W_t'' \cdot ({X^{res}})^T), \\
        A_t'' =\operatorname{softmax}(U_t'' \cdot O_t''), \\
        {X_t^{att}}''= {A_t}'' \cdot X^{res}, 
    \end{array}
\end{equation}
where $W_t'' \in\mathbb{R}^{d_a \times d_{res}} $ is the intermediate parameter matrix. 
$d_a$ is a hyperparameter, $O_t'' \in \mathbb{R}^{d_a \times n}$ is the intermediate result matrix and  $U_t''\in \mathbb{R}^{|L^t| \times d_a}$ is the code-specific attention parameter matrix. 
Finally, ${X_t^{att}}''\in \mathbb{R}^{|L^t|\times d_{res}}$ denotes code-specific document representation.

With the above description, we apply $R^t =\operatorname{concat}({X_t^{att}}',{X_t^{att}}'')\in \mathbb{R}^{|L^t|\times 2d_{res}}$ as the output of the MAU.

\subsubsection{Code Predicting Unit.}
For the level-$t$, 
we combine the result $R^t$ of MAU with the inherited 
information $c^{t-1}$ of the previous level to assign diagnosis codes to the patient. 
Specifically, the CPU uses a linear layer following a sigmoid transformation for each code:
\begin{equation}
    \label{eq13}
    \begin{array}{l}
        X_t^{cls} = \operatorname{concat} (\mathrm{broadcast}(c^{t-1}), R^t),\\
        \tilde{Y^{t}}=\sigma\left(X_t^{cls} \cdot W_y^{t} \right),
    \end{array}
\end{equation}
where $\operatorname{broadcast(\cdot)}$ is the process of making matrixes with different shapes have compatible shapes for arithmetic operations, 
$\sigma(\cdot)$ denotes an activation function, such as the $\operatorname{sigmoid}(x)=\frac{1}{1+e^{-x}}$,
$W_y^{t}\in\mathbb{R}^{(2d_{res}+d_c^{t-1}) \times 1 }$ is the parameter of the CPU, 
and $\tilde{Y^{t}} \in \mathbb{R}^{|L^t| \times 1}$ is the prediction results of the level-$t$.

\subsubsection{Dependency Passing Unit.}
We aim to preserve important information while reducing the harm caused by the previous 
level's error transmission. 
Therefore, we employ the combination of a linear layer and sigmoid function to imitate the 
gating mechanism to filter and integrate information as follows:

\begin{equation}
    \label{eq14}
    \begin{array}{c}
        Z=\operatorname{concat} ((\tilde{Y^{t}})^T, c^{t-1}), \\
        c^t = \sigma (Z \cdot W^t_{dpu}),
    \end{array}
\end{equation}
where $Z\in \mathbb{R}^{1 \times (|L^t|+d_c^{t-1})}$ 
and $W^t_{dpu} \in\mathbb{R}^{(|L^t|+d_c^{t-1}) \times d_c^t}$ is the parameter matrix.
Then, we can get the inter-level dependence $c^t\in\mathbb{R}^{1 \times d_{c}^t}$ based on the previous level's memory information 
and the prediction results of the current level. 

\subsection{Training}
For training, we combine all levels of multi-label binary cross-entropy as the loss:
\begin{equation}
    loss = \sum_t^{\mathcal{T}}{loss^t}
    =\sum_t^{\mathcal{T}}{\sum_{i=1}^{L^t}\left[-y_{i} \log \left(\tilde{y}_{i}\right)-\left(1-y_{i}\right) \log \left(1-\tilde{y}_{i}\right)\right], \text{where }\tilde{y}_{i} \in \tilde{Y^t}},
\end{equation}
where $loss_t$ indicates the loss function of level-$t$.

\section{Experiments}

\subsection{Dataset and Evaluation Metrics}
In this paper, we conduct experiments on a real-world dataset: the MIMIC-III dataset, 
which is widely used in clinical automatic diagnosis.
Following previous studies~\cite{mullenbach2018explainable,li2020icd}, 
we use the discharge summaries as the model's input and use the full codes and the top 50 most common codes for experiments. 
Specifically, for the  MIMIC-III full setting, it includes the 8,925 codes, 47,719, 1,631, and 3,372 discharge summaries used for training, validation, and testing, respectively. For the MIMIC-III top-50 setting, it includes 8,067, 1,574, and 1,730 discharge summaries used for training, validation, and testing, respectively. 
In addition, we expand the codes from fine to coarse according to the hierarchical inheritance structure of ICD because EHR data only have the finest-grained codes (i.e.the level-4 in Table~\ref{level_tablel}).
The specific statistical results are shown in Table~\ref{level_tablel}.

The evaluation metrics used in the experiments are Precision@K(K=5, 8, and 15), Macro-F1, Micro-F1, Macro-AUC and Micro-AUC.

\begin{table*}[htbp]
    \setlength{\abovecaptionskip}{0pt}
\setlength{\belowcaptionskip}{10pt}
    \renewcommand\arraystretch{1.0}
    \normalsize 
    \centering
    \caption{The statistics of hierarchical levels.}
    \label{level_tablel}
    \setlength{\tabcolsep}{2.2mm}{
        \scalebox{0.75}{
            \begin{tabular}{cccc|cccccccccc}
                \toprule
                Statistics                                 & full  & top-50      \\ 
                \midrule
                \# codes in level-1                        & 199   & 25     \\ 
                \# codes in level-2                        & 1,175  & 40     \\ 
                \# codes in level-3                        & 5,125  & 48      \\ 
                \# codes in level-4                        & 8,925  & 50      \\ 
                \# avg codes per  EHR in level-1           & 11.02 & 4.70    \\ 
                \# avg codes per  EHR in level-2           & 13.75 & 5.37    \\ 
                \# avg codes per  EHR in level-3           & 15.30 & 5.71    \\ 
                \# avg codes per  EHR in level-4           & 15.86 & 5.77   \\ 
                \bottomrule
            \end{tabular}
        }
    }  
\end{table*}

\subsection{Implementation Details}
We utilize PyTorch~\cite{paszke2019pytorch} to implement IHCE model and train it on a server with 4×V100 GPU. 
For the training setting, we use AdamW~\cite{loshchilov2017decoupled} for learning and set the learning rate and weight decay 
to 0.0001 and 0.00005, respectively. 
We set the dropout probability 0.4 and set the batch size to 16.
We also apply an early stop mechanism, 
in which the training will stop if the Micro-F1 score on the validation set does not improve in 10 continuous epochs.
Since our model has a number of hyperparameters, 
it is infeasible to search optimal values for all hyperparameters.
We keep the hyperparameters of the Feature Extraction  Module consistent with Li\cite{li2020icd}.
Specifically, the word embedding dimension $d_e$=100, 
the number of convolution kernels $m$ in feature extraction is 6, 
and the size of the convolution kernels $s_1,s_2,...s_m$ are set to ``3,5,9,15,19,25", $d_c=d_e$ and $d_r$=50.
Besides, we pre-train word embeddings on all the text in the training set using the word2vec~\cite{mikolov2013efficient} implemented by gensim~\footnote{\scriptsize https://radimrehurek.com/gensim/}.
The maximum length of a token sequence is 2,500, and the one that exceeds this length will be truncated.
For the remaining parameters, we use the grid to search for the optimal hyperparameters. 
Specifically, we set the number of hidden layers to 1, and the hidden layer size $d_g=300$ for GCN.
In addition, we set $d_a$=300 for ORL's attention dimension, and $d^t_c=500\text{ } (t=1,2,...,\mathcal{T}-1)$ for all DPUs' parameters dimension.

\subsection{Baselines}
We compared IHCE with the following baselines, including machine learning and deep learning models:
\begin{itemize}
    \item \textbf{LR:} which is a bag-of-words logistic regression model.
    \item \textbf{H-SVM~\cite{perotte2014diagnosis}:} which designs a hierarchical SVM algorithm from root to leaf node by utilizing the hierarchical structure of diagnosis codes.
    \item \textbf{Bi-GRU~\cite{mullenbach2018explainable}:} which employs bidirectional gated recurrent units to learn clinical note's representation for automatic diagnosis task.
    \item \textbf{C-MemNN~\cite{prakash2016condensed}:} which combines the memory network with iterative compression memory representation to improve diagnosis accuracy.
    \item \textbf{C-LSTM-Att~\cite{shi2017towards}:} which uses an LSTM-based language model to generate clinical note and diagnosis code representations as well as an attention mechanism to resolve the mismatch between notes and codes.
    \item \textbf{LEAM~\cite{wang2018joint}:} which is proposed for text classification task by projecting labels and words in the same embedding space and using the cosine similarity to predict the label of text.
    \item \textbf{HARNNN~\cite{huang2019hierarchical}} which is initially used for multi-label text classification and considers the hierarchy of categories. We apply it to the automatic diagnosis.
    \item \textbf{CNN~\cite{mullenbach2018explainable}:} which uses a single layer convolutional neural network and a max-pooling layer for automatic diagnosis task.
    \item \textbf{CAML and DR-CAML~\cite{mullenbach2018explainable}:} which assign diagnosis codes based on clinical text by using CNN to aggregate information among the clinical note and attention mechanism to select the most relevant segment for each possible code. DR-CAML further uses text description as a regularization.
    \item \textbf{MultiResCNN~\cite{li2020icd}:} which utilizes multi-fliter convolutional neural networks and residual networks for automatic diagnosis and becomes the SOTA model on MIMIC-III.

\end{itemize}

\subsection{Overall Performance}
In this section, we compare the IHCE with existing works for clinical automatic diagnosis.
Table~\ref{overall_mimic} shows our overall performance on MIMIC-III full setting and MIMIC-III 50 setting.
$\mathcal{T}=3$ means that our experiment is based on the last three levels (i.e., level-$2$ to level-$4$ in Table \ref{level_tablel}) in the hierarchy. 
Our model IHCE surpasses all baselines on both settings.
The results indicate that IHCE is able to effectively perform clinical automatic diagnosis by exploiting the hierarchy and co-occurrence structure of the medical ontology and the attention mechanism.
The specific analysis is as follows:
\begin{table*}[htbp]
\setlength{\abovecaptionskip}{0pt}
\setlength{\belowcaptionskip}{10pt}
    \renewcommand\arraystretch{1.1}
    \normalsize 
    \centering
    \caption{Overall performance on MIMIC-III, where ``-" means that the baseline did not report the result of the corresponding metric.}
    \label{overall_mimic}
    \setlength{\tabcolsep}{0.9mm}{
        \scalebox{0.85}{
            \begin{threeparttable}
                \begin{tabular}{c|cc|cc|cc|cc|cc|cc|c}
                    \hline    
                    \toprule
                    \multirow{3}{*}{Models}  &\multicolumn{6}{c|}{MIMIC-III full} & \multicolumn{5}{c}{MIMIC-III top-50} \\\cmidrule(r){2-12}
                    
                    &\multicolumn{2}{c}{AUC} & \multicolumn{2}{c}{F1-score} &\multicolumn{2}{c|}{P@K} & \multicolumn{2}{c}{AUC} & \multicolumn{2}{c}{F1-score} &\multicolumn{1}{c}{P@K} \\
                    \cmidrule(r){2-3} \cmidrule(r){4-5} \cmidrule(r){6-7} \cmidrule(r){8-9} \cmidrule(r){10-11} \cmidrule(r){12-12}
                & Macro & Micro & Macro & Micro  &8 &15 & Macro & Micro & Macro & Micro  &5   \\  
                    \midrule
                    LR  & 56.1 & 93.7 & 1.1 & 27.2 & 54.2& 41.1& 82.9& 86.4& 47.7& 53.3& 54.6 \\
                    H-SVM & - & - & - & 44.1 & - & -& -& -& -& -& -  \\
                    \midrule
                    C-MemNN  & - & -& -& -& -& -& 83.3 & - & - & - & 42.0  \\
                    C-LSTM-Att   & - & -& -& -& -& -&- & 90.0 & - & 53.2 & -   \\
                    HARNN & - & - & - & 40.5 & - & - & - & - & - & - & - \\  
                    BiGRU & 82.2 & 97.1 & 3.8 & 41.7 & 58.5 & 44.5 & 82.8 & 86.8 & 48.4 & 54.9 & 59.1  \\
                    LEAM  & - & -& -& -& -&-& 88.1 & 91.2 & 54.0 & 61.9 & 61.2  \\    
                    CNN & 80.6 & 96.9 & 4.2 & 41.9 & 58.1 &44.3 & 87.6 &90.7 &57.6 &62.5 &62.0 \\
                    CAML & 89.5 & 98.6 & 8.8 & 53.9  & 70.9 & 56.1 & 87.5 & 90.9 & 53.2 & 61.4 & 60.9 \\
                    DR-CAML & 89.7 & 98.5 & 8.6 & 52.9 & 69.0 & 54.8  & 88.4 & 91.6 & 57.6 & 63.3 & 61.8  \\
                    MultiResCNN & 91.0 & 98.6 & 8.5 & 55.2 & 73.4 & 58.4& 89.9 & 92.8 & 60.6 & 67.0 & 64.1   \\
                    \midrule
                    {\bf IHCE($\mathcal{T}=3)$}&{\bf 92.9 } &{\bf 98.9} &{\bf  10.4} &{\bf  57.3} &{\bf  73.5} &{\bf  58.7} &{\bf 91.0 } &{\bf 93.6} &{\bf  64.7} &{\bf  69.6} &{\bf  65.2}\\
                    \bottomrule
                    \hline
                \end{tabular}
            \end{threeparttable}
        }
    }  
\end{table*}

(1) In the MIMIC-III full setting, compared with the SOTA method MultiResCNN, 
the IHCE improves Macro-AUC, Macro-F1 and Micro-F1 by 2.1\%, 22.4\% and 3.8\%, respectively.
It is worth noting that all models have low Macro-F1 scores on MIMIC-III full setting because the diagnosis codes space is too large, 
and the distribution is extremely unbalanced. 
Nevertheless, what is exciting is that our model has \textbf{18.2\%} and \textbf{22.4\%} improvements in this metric compared to CAML and MultiReCNN, respectively. 
The reason is the IHCE considers hierarchical inheritance structure and dependencies. 
So the IHCE can assists the processing of low-frequency codes based on high-level prediction results. 
Similarly, we can observe that H-SVM with a hierarchical structure is better than BiGRU without a hierarchical structure in Micro-F1.
However, the performance of H-SVM is lower than that of CAML and MultiReCNN because CAML and MultiReCNN utilize a primary attention mechanism to improve the ability to retrieve critical information. Furthermore, compared to CAML and MultiResCNN, our model has multiple attention mechanisms, so our model has more robust key information retrieval capabilities and surpasses them in all metrics.

(2) In the MIMIC-III top-50 setting, compared with the SOTA method MultiResCNN,
the IHCE improves Macro-F1 and Micro-F1 by 6.8\% and 3.9\%, respectively. 
Although there are only 50 diagnosis codes in MIMIC-III top-50 setting, 
it still shows a slight long-tail effect. 
The IHCE has a significant improvement on the Macro-f1, indicating that our model can employ the hierarchical structure to alleviate this problem. It is worth noting that even though DR-CAML utilize codes description as regularization to assist in the allocation of diagnosis codes that are difficult to predict, the effect is still limited compared to CNN. However, the IHCE utilizes the co-occurrence structure between codes to solve this problem better.

\subsection{Ablation Study}
In this section, to verify each component's effectiveness in the IHCE, 
we perform ablation studies. The specific results are shown in Table \ref{Ablation}.
It is observed that removing each component will cause F1 to decrease, 
which illustrates the effectiveness of each component of our model. 
(1) \textbf{HPL's effectiveness:} After removing the HPL module, 
the macro-average metrics drop significantly, 
indicating that the inheritance-guided hierarchical assignment mechanism introduced by our IHCE 
has a significant effect on solving the long-tail effect.
(2) \textbf{ORL's effectiveness:} After ORL is removed, 
the overall performance of IHCE declines because the method cannot model disease co-occurrence relationships. 
However, this ability is beneficial for assigning diseases for which it is not easy to find textual clues in the clinical note.
(3) \textbf{Attention mechanism's effectiveness:} 
We only retain the \textsl{Code Specific Attention} module, which expands the attention mechanism in MultiResCNN and improves almost all metrics. 
It shows that our attention mechanism can better extract essential information to prevent the situation of finding a needle in a haystack.

\begin{table*}[htbp]
\setlength{\abovecaptionskip}{0pt}
\setlength{\belowcaptionskip}{10pt}
    \renewcommand\arraystretch{1.0}
    \normalsize 
    \centering

    \caption{Ablation study results, where “w/o” indicates without.}
    \label{Ablation}
    \setlength{\tabcolsep}{2.2mm}{
        \scalebox{0.75}{
            \begin{tabular}{cccc|cccccccccc}
                \hline    
                \toprule
                \multirow{2}{*}{Models} & \multicolumn{3}{c}{MIMIC-III full} & \multicolumn{3}{c}{MIMIC-III top-50}  \\
                \cmidrule(r){2-4} \cmidrule(r){5-7} 
                & Macro-AUC & Macro-F1 & Micro-F1  & Macro-AUC & Macro-F1 & Micro-F1\\ 
                \midrule
                MultiResCNN(SOTA) & 91.0 & 8.5 & 55.2 & 89.9  & 60.6 & 67.0 \\
                \midrule 
                {\bf w/o ORL\&HPL} & 91.0 & 8.7 &55.9 & 89.9 & 61.2 &66.9 \\
                {\bf w/o HPL}      & 92.6 & 9.2 &56.0 & 89.9 & 62.1 &67.5 \\
                {\bf w/o ORL}    & {\bf 93.1} & 10.0 & 56.7  &90.6 &63.6 &68.5 \\
                \midrule
                {\bf IHCE($\mathcal{T}=3$)} &92.9 &{\bf  10.4} &{\bf  57.3} &{\bf 91.0 } &{\bf  64.7} &{\bf  69.6} \\
                \bottomrule
                \hline
            \end{tabular}
        }
    }  

\end{table*}

\subsection{Performance at Different Levels}
In the clinical automatic diagnosis task, it is important to assign the diagnosis codes of the last level to the patient. 
It is also essential to evaluate the performance at different levels because, in some cases, a different granularity of codes may be required.
\begin{figure}
    \includegraphics[scale=0.3]{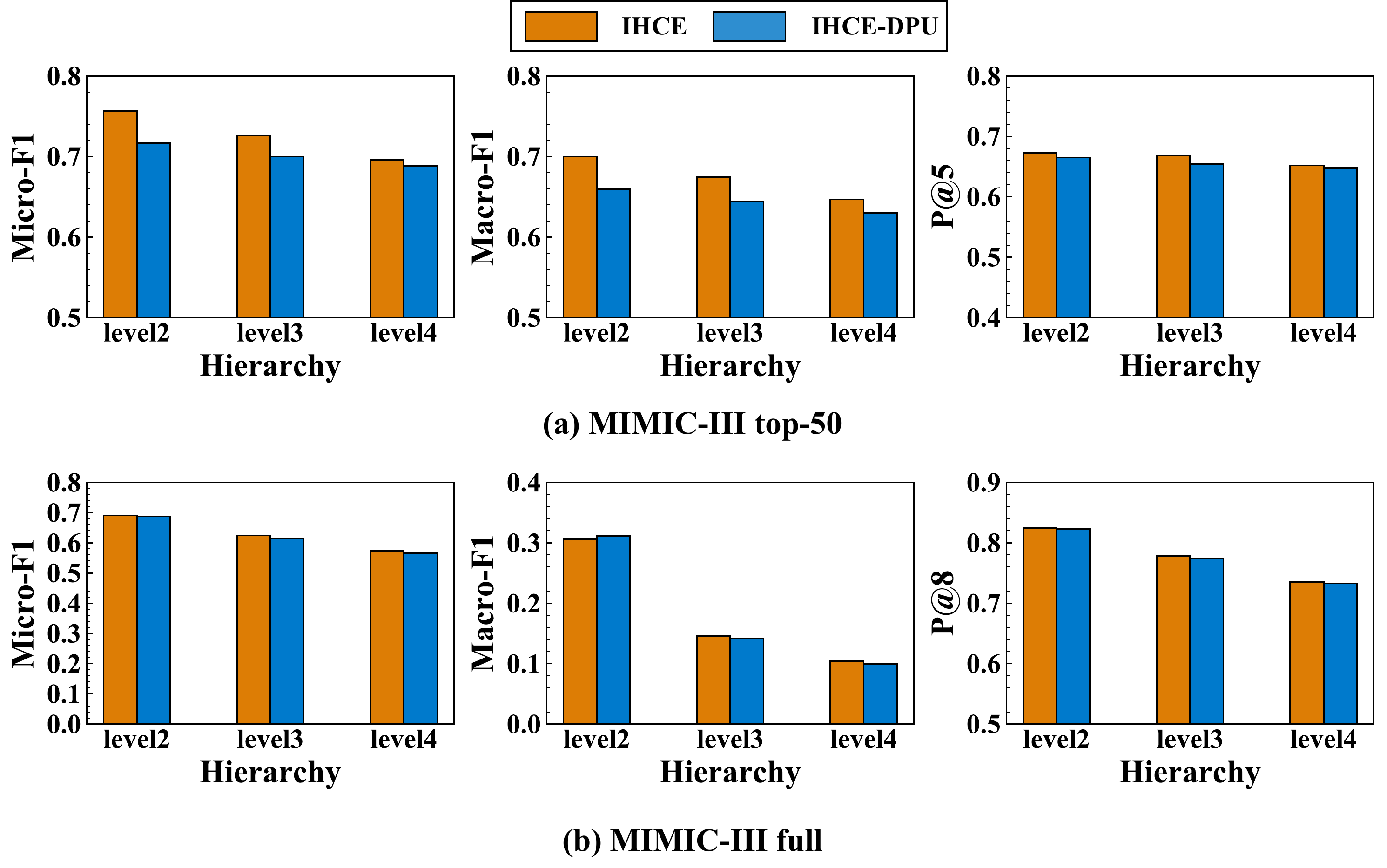}
    \centering
    \caption{Performance at different levels in hierarchy.} \label{levels_result}
\end{figure}
Therefore, we compared the performance of IHCE and IHCE-DPU at each hierarchical level. 
Note that this comparison is based on $\mathcal{T}=3$.
The IHCE-DPU ignores the dependency between the levels by removing the DPU in the HPM.
In Fig. \ref{levels_result}, we can see that the performance of IHCE at almost all levels is better than IHCE-DPU. 
Moreover, we can also notice that the performance on all metrics tend to
decrease when the hierarchy deepens, and the trend on Macro-F1 in MIMIC-III full setting is the most obvious.
The reason is that as the level deepens, the number of codes of this level will increase rapidly
(e.g., the MIMIC-III full setting has 5,125, 8,925 unique codes in level-3 and level-4 respectively, 
as shown in Table \ref{level_tablel}).
Moreover, we can notice that IHCE reduces this negative factor compared with IHCE-DPU by modeling the dependency among different hierarchical levels.

\subsection{Effect of the Number of Hierarchical Levels}
In this section, we turn to figure out the effect of the number of hierarchical levels, i.e., $\mathcal{T}$. 
To that end, a series of experiments are conducted to evaluate the effectiveness under different settings. 
Specifically, $\mathcal{T}=n$ means choosing the last $n$ levels in Table \ref{level_tablel}. 
For example, $\mathcal{T}=2$ means that we choose level-3 and level-4.
\begin{figure}
    \includegraphics[scale=0.4]{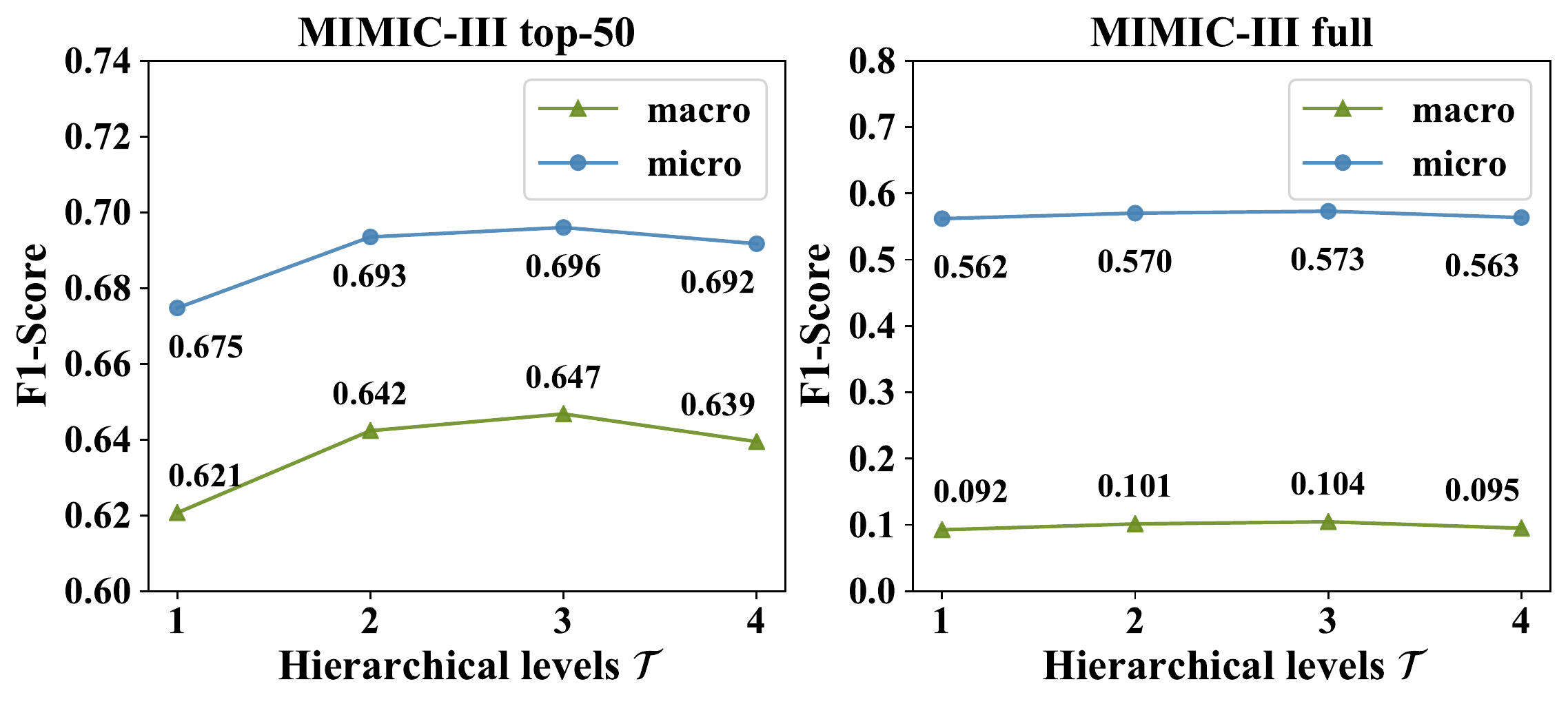}
    \centering
    \caption{Performance by varying the number of hierarchical levels.} \label{hie_levels}
\end{figure}
From Fig. \ref{hie_levels}, 
we can conclude that the models that consider hierarchical structure preform much better than models that do not.
The performance rises when the number $\mathcal{T}$ of levels increases because high-level information has a guiding effect on the low level. However, the performance decreases when the  $\mathcal{T}$ continuously increases. 
The reason is that when the number of codes between different levels is not an order of magnitude, errors caused by high-level results will still seriously affect low-level levels, although DPU has a mitigating effect.
Specifically, for the MIMIC-III full setting, when $\mathcal{T}$=4, the model will extend level-1 with only 199 diagnosis codes, which is not in the same order of magnitude as other levels.
For the MIMIC-III top-50 setting, 
each level's magnitude is not much different, 
and the impact of this error will also be reduced.
\section{Conclusion}
In this paper, we proposed a novel Inheritance-guided Hierarchical Assignment with Co-occurrence-based Enhancement (IHCE) framework for clinical automatic diagnosis, which could jointly exploit code hierarchy and code co-occurrence. 
We utilized GCN to obtain the correlation between medical ontology. 
Moreover, we proposed a hierarchical joint prediction strategy based on the attention mechanism.
Experimental results on real medical datasets show that our model has obtained 
state-of-the-art performance with substantial improvements in different evaluation metrics.
We believe that our method can also be used for other tasks that require the application of hierarchical structure and label co-occurrence, such as hierarchical multi-label classification.

\subsubsection{Acknowledgements}
This research was partially supported by grants from the National Key Research and Development Program of China (Grant No.2018YFB1402600), the National Natural Science Foundation of China (Grant No.62072423), and the Key Research and Development Program of Anhui Province (No.1804b06020377).
\bibliographystyle{splncs04}

\bibliography{ref}

\end{document}